\documentclass[
twocolumn,
]{ceurart}

\usepackage{listings}
\usepackage{textcomp}

\usepackage{amsmath}
\usepackage{amsthm}
\usepackage{multirow}
\usepackage{adjustbox}
\usepackage{booktabs}
\usepackage{graphicx}
\usepackage[shrink]{cuted}
\usepackage{algorithm}
\usepackage{algorithmic}
\usepackage{tabularx}
\usepackage{graphicx}
\usepackage{subcaption}
\usepackage{float}
\usepackage{xcolor}
\usepackage[utf8]{inputenc}
\usepackage{hyperref}
\usepackage[switch]{lineno}
\usepackage{svg}
\usepackage{epsfig}
\begin{document}

\copyrightyear{2024}
\conference{ConfWS'24: 26th International Workshop on Configuration, Sep 2--3, 2024, Girona, Spain}

\conference{ConfWS'24: 26th International Workshop on Configuration, Sep 2--3, 2024, Girona, Spain}

\title{Developing an Algorithm Selector for Green Configuration in Scheduling Problems}


\author[1]{Carlos March}[orcid=0009-0009-7525-9133,email=cmarmoy@upv.es,url=https://gps.blogs.upv.es/,]
\cormark[1]
\fnmark[1]
\author[1,2]{Christian Pérez}[orcid=0000-0002-9121-7939,email=cripeber@upv.es,url=https://gps.blogs.upv.es/,]
\fnmark[1]
\author[1]{Miguel Salido}[orcid=0000-0002-4835-4057,email=msalido@upv.es,url=https://gps.blogs.upv.es/,]
\address[1]{Universitat Politècnica de València, Instituto de Automática e Informática Industrial, Camì de Vera S/N, Valencia, Spain}
\address[2]{Universitat Politècnica de València, valgrAI - Valencian Graduate School and Research Network of Artificial Intelligence, Camì de Vera S/N, Valencia, Spain}

\cortext[1]{Corresponding author.}
\fntext[1]{These authors contributed equally.}

\begin{abstract}
The Job Shop Scheduling Problem (JSP) is central to operations research, primarily optimizing energy efficiency due to its profound environmental and economic implications. Efficient scheduling enhances production metrics and mitigates energy consumption, thus effectively balancing productivity and sustainability objectives. Given the intricate and diverse nature of JSP instances, along with the array of algorithms developed to tackle these challenges, an intelligent algorithm selection tool becomes paramount.
This paper introduces a framework designed to identify key problem features that characterize its complexity and guide the selection of suitable algorithms. Leveraging machine learning techniques, particularly XGBoost, the framework recommends optimal solvers such as GUROBI, CPLEX, and GECODE for efficient JSP scheduling. GUROBI excels with smaller instances, while GECODE demonstrates robust scalability for complex scenarios.
The proposed algorithm selector achieves an accuracy of 84.51\% in recommending the best algorithm for solving new JSP instances, highlighting its efficacy in algorithm selection. By refining feature extraction methodologies, the framework aims to broaden its applicability across diverse JSP scenarios, thereby advancing efficiency and sustainability in manufacturing logistics.
\end{abstract}

\maketitle

\section{Introduction}
The Job Shop Scheduling Problem (JSP) is a cornerstone issue in operations research and optimization, serving as a critical benchmark for assessing the performance of various algorithms. JSP entails the complex task of scheduling jobs on machines in a manufacturing environment to optimize several performance metrics, such as makespan, flow time, tardiness, resource utilization, and energy consumption \cite{xiong2022survey}. Effective benchmarking of JSP solutions requires a multi-faceted evaluation of these metrics, particularly focusing on makespan, energy consumption, and tardiness to gauge scheduling efficiency and resource utilization \cite{fontes2023hybrid}. Tools like JSPLIB play a vital role in these benchmarking efforts by providing researchers with diverse instances derived from significant studies and experiments, thereby enhancing the evaluation of algorithms \cite{torres2014psplib}.

Understanding the characteristics of problem instances is essential for effective benchmarking in JSP. Critical factors include the number of jobs and machines, variability in processing times, machine availability, and precedence relationships, all of which significantly impact algorithm performance \cite{el_kholany2022problem}. Additionally, considering energy consumption, which varies based on machine speed and operational factors, adds another layer of complexity \cite{perez2020metaheuristic}. Achieving a balance between energy consumption and scheduling decisions is crucial for attaining energy efficiency without compromising production goals \cite{dai2019multi}.

JSP's focus on energy efficiency has intensified in recent years due to its substantial environmental and economic impacts \cite{perez2023hybrid}. Researchers have investigated strategies such as employing speed-adjustable machines and vehicles to minimize energy consumption while maintaining productivity \cite{salido2016genetic}. Advanced algorithms and optimization techniques have been developed to address these energy-related challenges, taking into account factors like machine speed, idle time, and energy requirements \cite{jyothi2023minimizing}. Real-world implementations of these strategies have demonstrated tangible benefits, including cost savings and positive environmental effects \cite{ham2021energy}.

In addition to traditional optimization methods, machine learning techniques are increasingly being utilized to recommend algorithms for solving problems within the JSP family. For instance, \citeauthor{muller_algorithm_2022} designed a system capable of selecting the most suitable solver for addressing Flexible JSP by leveraging machine learning approaches \cite{muller_algorithm_2022}. Similarly, \citeauthor{strassl_instance_2022} \cite{strassl_instance_2022} analyzed JSP instances without energy consumption from the literature to extract features that inform algorithm performance, resulting in a homogeneous set of instances with consistent characteristics. These features were then used to train various models, with Random Forest achieving the highest accuracy at 90\% \cite{strassl_instance_2022}.

In conclusion, the integration of machine learning techniques into JSP research provides new avenues for improving algorithm selection and performance, particularly in handling complex and varied instances. This integration enhances the efficiency and effectiveness of job shop scheduling by combining the strengths of traditional optimization approaches with innovative machine learning methods. The ongoing advancements in this field are driving both academic research and practical applications toward more sustainable and innovative solutions.

\section{Problem Description and Model Formulation}
The JSP tackled in this study emphasizes its intricate energy considerations. The JSP poses a significant computational challenge, being NP-complete due to its difficulty finding optimal solutions within reasonable time frames.

The core challenge of the JSP involves optimizing task allocation across multiple jobs and machines while minimizing key criteria, notably the total job completion time. However, achieving this optimization is complex due to various real-world constraints and dependencies, contributing to the JSP's NP-completeness. The combinatorial explosion of possible job and machine combinations further complicates the problem, making exhaustive exploration impractical as the number of jobs and machines grows.

\subsection{Mixed Integer Programming}
The JSP involves various sets, parameters, variables, and constraints crucial for formulation and solution:

\emph{Sets}:
\begin{itemize}
	\item $J=\{1,\dots,n\}$, the set of jobs.
	\item $M=\{1,\dots,m\}$, the set of machines.
	\item $S=\{1,\dots,s\}$, the set of speeds.
	\item $T_{j}$,~$\forall j\in J$, the set of tasks in job $j$. In standard JSP $T_{j} = M$.
 \end{itemize}

\emph{Parameters}:
\begin{itemize}
	\item $D_{jt}$,~$\forall j \in J$,~$\forall t \in T_j$, the due date of task job $t_{jt}$.
	\item $R_{jt}$,~$\forall j \in J$,~$\forall t \in T_j$, the release date of task job $t_{jt}$.
	\item $P_{jts}$,~$\forall j \in J$,~$\forall t \in T_j$, the processing time of task job $t_{jt}$ on machine $t$ with speed $s$.
	\item $E_{jts}$,~$\forall j \in J$,~$\forall t \in T_j$, the energy consumption for processing task job $t_{jt}$ with speed $s$.
\end{itemize}

\emph{Variables}:
\begin{itemize}
	\item $c_{jt}$,~$\forall j\in J,$~$\forall t\in T_j$, the completion time of task job $t_{jt}$
	\item $tt_{jt}$,~$\forall j\in J,$~$\forall t\in T_j$, tardiness of task job $t_{jt}$ with respect to its due date
	\item $x_{mjts}\in\{0,1\}$,~$\forall m\in M$,~$\forall j\in J$,~$t \in T_{j}$, binary sequencing variables (i.e., $x_{mjts} = 1$ denotes that task $t$ of job $j$ is performed with speed $s$ on machine $m$)
	\item $ y_{mijpq} \in\{0,1\}$,~$\forall m\in M$,~$\forall i,j\in J$,~$\forall p,q\in T_i, T_j$,~$i \neq j$, binary assignment variables (i.e., $y_{ijpq} = 1$ denotes that task $p$ of job $i$ precedes task $q$ of job $j$ on machine $m$)
\end{itemize}

\begin{equation}
    \label{eq:fo}
    \phi^* = \operatorname*{arg\,min}_{\phi \;\in\; \Phi}\, [MK(\phi),EC(\phi),TT(\phi)] 
\end{equation}

subject to:

\begin{flalign}
    &\sum_{m\in M} x_{mjts} = 1  && \\ 
    &&\forall j \in J, \;\forall t \in T_j \; \forall s \in S &\nonumber
\end{flalign}
\begin{flalign}
    &\sum_{m\in M} y_{mijpq} = 1 & \\
    &&\forall i,j \in J, \; \forall p, q \in T_i,T_j, &\nonumber\\
    && i \neq j , p \leq q  &\nonumber
\end{flalign}
\begin{flalign}
    &tt_{mjt} \geq c_{mjt} - D_{jt}   && \\
    &&\forall m\in M, \forall j \in J, &\nonumber\\
    &&\forall t\in T_j,\; x_{mjt} = 1 &\nonumber
\end{flalign}
\begin{flalign}
    &c_{mjt} \geq R_{jt} + P_{mjts}  && \\
    &&\forall m\in M, \; \forall j \in J, &\nonumber\\
    && \forall t\in T_j ,\; \forall s \in S, \; x_{mjts} = 1 &\nonumber
\end{flalign}
\begin{flalign}
    &c_{mjs}\geq  c_{mip} + P_{mips} && \\
    &&\forall m\in M,\; \forall i,j\in J, &\nonumber\\
    &&\forall p ,q \in  T_i ,T_j, \; \forall s \in S, &\nonumber\\
    && i \neq j \; \wedge \; p < q \; \wedge y_{mijps} = 1   &\nonumber
\end{flalign}
\begin{flalign}
    &c_{mjt}\geq 0 \; , t_{mjt}\geq 0  && \\
    &&\forall m\in M, \forall j\in J \forall t\in T_j &\nonumber 
\end{flalign}
This model seeks the optimal solution $\phi^*$ that minimizes the three measures mentioned in equation \ref{eq:fo}. considering the constraints associated: the maximum makespan of all task jobs $MK(\phi)$, the total energy consumption $EC(\phi)$, and the total tardiness $TT(\phi)$. The simultaneous optimization of these objectives requires a delicate balance between the various considerations and constraints of the problem. Therefore, two approaches to optimizing the problem solutions are proposed, allowing us to analyze the methods' behavior better.

\subsection{Mono-objective optimization}
This section presents the mono-objective optimization for a specific scheduling problem involving multiple jobs and machines, emphasizing key performance metrics such as makespan, energy consumption, and total tardiness.

\begin{flalign}
    \label{eq:FM}
    &f^m  = \max_{\substack{j \in J\\ m \in M}}(c_{jm}) \\
    \label{eq:FE}
    &f^e  = \sum_{j\in J}\sum_{t\in T_j} E_{jt} \\
    \label{eq:FTT}
    &f^{tt} = \sum_{m \in M}\sum_{j\in J}\sum_{\substack{t\in T_j}} tt_{mjt}
\end{flalign}

Equation \ref{eq:FM} represents the makespan, which is the maximum completion time among all machines, by calculating the total processing time of all job tasks on each machine and selecting the maximum value across all machines. Equation \ref{eq:FE} describes energy consumption by computing the total energy consumed by all job tasks across all machines. Lastly, Equation \ref{eq:FTT} is formulated to show the total tardiness, which represents the number of time units of each job or operation that are performed outside its time window, i.e., the period of time between the release date and the due date.

\begin{flalign}
    \label{eq:obmulti}
    min~~ \frac{\displaystyle f^m - m^{-}_{1}}{m^{+}_{1} - m^{-}_{1}} ~~+ 
    \frac{\displaystyle f^e - m^{-}_{2}}{m^{+}_{2} - m^{-}_{2}} ~~+ 
    \frac{\displaystyle f^{tt}}{m^{+}_{1}}
\end{flalign}

Minimizing the objective Function \ref{eq:obmulti} aims to find a solution that achieves a balanced trade-off among the components. The values $m_{{1,2}}^{+}$ and $m_{{1,2}}^{-}$ are used to normalize the $\phi^*$ solution obtained in the three-dimensional objective space. This allows a correct comparison between the values of the objective function in minimizing the problem, giving the same weight to all the parts, and avoiding any of the variables dominating the search.

\begin{flalign}
    \label{eq:minmaxmakespan}
     \displaystyle m^{+}_{1} =& \sum_{j\in J}(\sum_{m \in M} \max_{s \in S}(P_{jms})) \\
     \displaystyle m^{+}_{2} =& \sum_{m \in M}(\sum_{j\in J} \max_{s \in S}(E_{jms})) \\ 
     \displaystyle m^{-}_{1} =& \max_{j \in J}(\sum_{m \in M} \min_{s \in S}(P_{jms})) \\
     \displaystyle m^{-}_{2} =& \sum_{m \in M}(\sum_{j\in J} \min_{s \in S}(E_{jms})) 
    \label{eq:minmaxenergy}
\end{flalign}

Equations between \ref{eq:minmaxmakespan} and \ref{eq:minmaxenergy} determine both maximum ($ m^{+}_{1} $ and $ m^{+}_{2} $) and minimum ($ m^{-}_{1} $ and $ m^{-}_{2} $) values for makespan and energy consumption respectively. For makespan, these values signify the maximum and minimum completion times across all machines, accounting for the maximum and minimum processing times of job tasks on each machine. Similarly, in terms of energy consumption, they represent the maximum and minimum energy utilized among all machines, considering the maximum and minimum energy consumption of all job tasks on each machine.

\section{Algorithm Selector}

The selection of algorithms for a given problem $ JSP $ involves identifying the most appropriate algorithm from a collection capable of solving $ JSP $, taking into account the specific characteristics of $ JSP $. Rubinoff \cite{rubinoff_algorithm_1976} formalized this process of algorithm selection. Rubinoff defined key elements, including the problem space $ X $, representing all instances of $ JSP $; the algorithm space $ A $, encompassing algorithms capable of solving any $ jsp \in X $; and a performance metric $ y $, which quantifies algorithm effectiveness for solving $ jsp \in X $.

The core objective is to establish a function $ S: X \rightarrow A $ that, for each problem instance $ jsp \in X $, selects the optimal algorithm from $ A $ based on metric $ y $. To effectively characterize each $ jsp $, a feature set $ F $ is constructed to represent $ p $ and assist in the decision-making process for $ S $. Consequently, $ S $ is defined as a composite function $ S = T \circ G $, where $ G: X \rightarrow \mathbb{R}^{|F|} $ maps $ p $ to its feature vector in $ \mathbb{R}^{|F|} $, and $ T: \mathbb{R}^{|F|} \rightarrow A $ selects the algorithm from $ A $ based on this feature representation. The choice of $ F $ is critical as it must be informative and accurately represent the characteristics of $ JSP $.

\begin{figure*}[hbtp]
    \centering
     \includegraphics[width=0.85\linewidth]{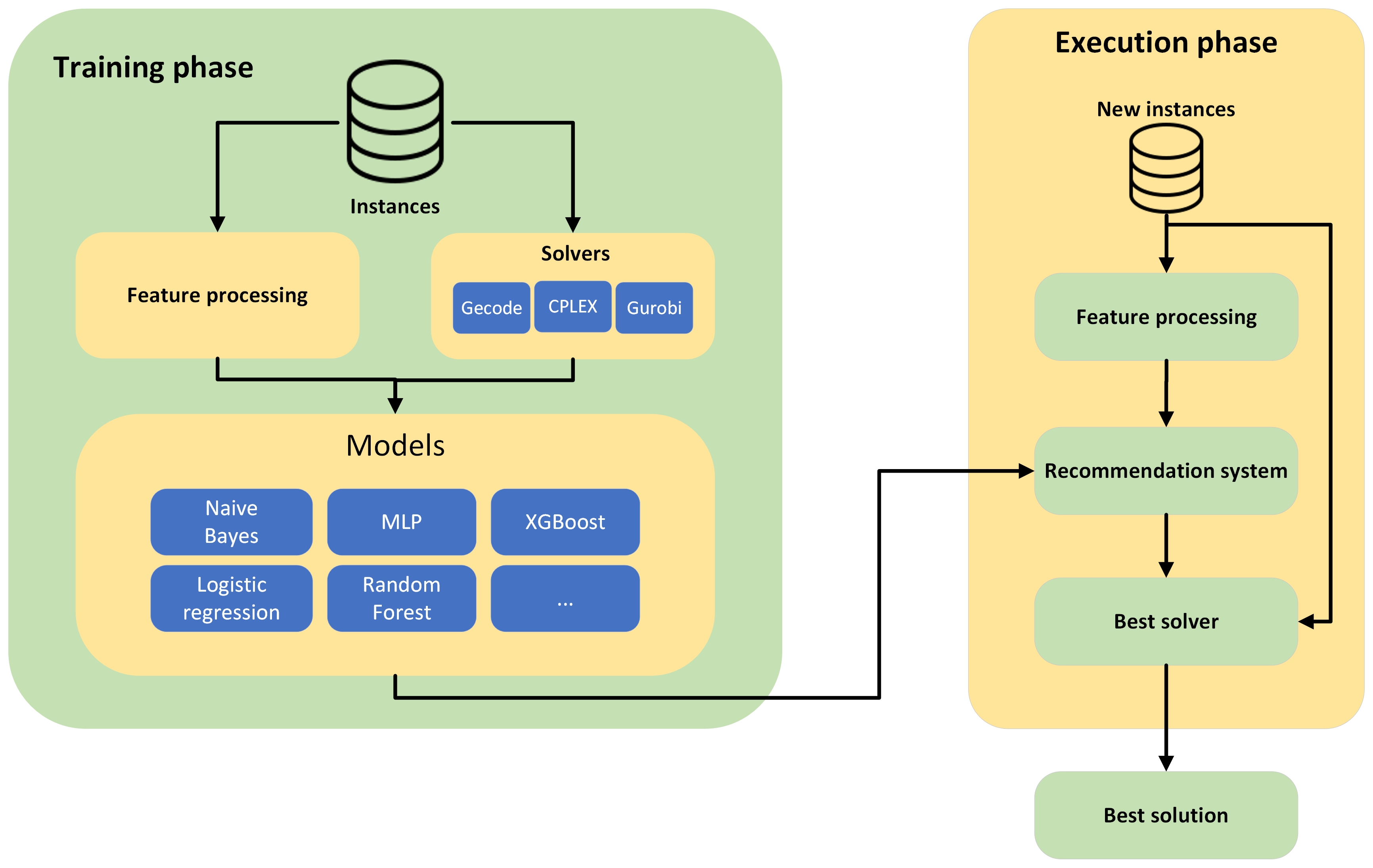}
    \caption{Structure of the proposed recommender system.}
    \label{fig:estrucRecomendador}
\end{figure*}

Considering the modeling of algorithm selectors in Figure \ref{fig:estrucRecomendador}, an algorithm selector structure is proposed, where it can be seen that it is composed of a training phase, in which the features of a set of instances are processed to generate a set of data and are solved using three solvers: GECODE, CPLEX, and GUROBI. Once the instances have been solved, the extracted features are related to the best algorithm that has solved that instance, and different machine learning models are trained in order to validate which is the one that obtains the best accuracy and thus use it to recommend future instances.

The following subsections detail each of the training processes.

\subsection{Feature processing}

For each instance, we extract the typical characteristics of a JSP (Job Shop Scheduling) problem, such as the number of jobs $|J|$, the number of machines $|M|$, the type of Release date, and Due date constraint $Rd/Dd$, and the number of speeds $|S|$. Additionally, we extract other features that are obtained in a less direct manner and aim to be as informative as possible about the complexity of the instance they represent. The extra features extracted are:

\begin{equation}
    \label{eq:maxTproc}
    \text{max}(P)
\end{equation}

\begin{equation}
    \label{eq:meanTproc}
    \text{mean}(P)
\end{equation}

\begin{equation}
    \label{eq:minTproc}
    \text{min}(P)
\end{equation}

The maximum processing time (\ref{eq:maxTproc}) represents the longest time required to complete any single operation within the job set. The mean processing time ( Equation \ref{eq:meanTproc}) gives the average duration of the operations, providing an overall sense of the job length. The minimum processing time (Equation \ref{eq:minTproc}) shows the shortest time needed for any operation, indicating the fastest job segment.


\begin{equation}
  \label{eq:maxEnergyValue}
  \max(E)
\end{equation}
\begin{equation}
  \label{eq:meanEnergy}
  \text{mean}(E)
\end{equation} 
\begin{equation}
  \label{eq:minEnergyValue}
  \min(E)
\end{equation}

The maximum energy consumption (Equation \ref{eq:maxEnergyValue}) indicates the highest energy required for any single operation. The mean energy consumption (Equation \ref{eq:meanEnergy}) provides the average energy used across all operations, reflecting the overall energy profile. The minimum energy consumption (Equation \ref{eq:minEnergyValue}) shows the lowest energy usage for an operation, highlighting the least energy-intensive job segment.

\begin{equation}
    \sum_{j \in J} \left(\sum_{m \in M} \max_{s \in S}(P_{jms})\right)
    \label{eq:maxMakespan}
\end{equation}

Maximum makespan (Equation \ref{eq:maxMakespan}) represents the maximum makespan of the instance obtained, assuming that all operations are performed serially with their maximum processing time. This value gives the longest possible duration to complete all jobs, assuming no parallel processing.

\begin{equation}
    \max_{j \in J} \left(\sum_{m \in M} \min_{s \in S}(P_{jms})\right)
    \label{eq:minMakespan}
\end{equation}

Minimum makespan (Equation \ref{eq:minMakespan}) represents the makespan of the solution obtained by considering that the operations can be performed in parallel and do not overlap. This makespan represents a lower bound of the possible makespan in a solution, indicating the shortest time to complete all jobs if perfectly parallelized.

\begin{equation}
    \sum_{m \in M} \left(\sum_{j \in J} \max_{s \in S}(E_{jms})\right)
    \label{eq:maxEnergy}
\end{equation}
\begin{equation}
    \sum_{m \in M} \left(\sum_{j \in J} \min_{s \in S}(E_{jms})\right)
    \label{eq:minEnergy}
\end{equation}

The sum of the maximum (Equation \ref{eq:maxEnergy}) and minimum (Equation \ref{eq:minEnergy}) energy consumption is obtained by adding for each operation its maximum and minimum energy consumption, respectively. These values provide insights into the total energy requirements of the job set under extreme conditions.

\begin{equation}
\begin{cases}
    -1 & Rd/Dd = 0 \\
    \displaystyle\sum_{j \in J} \left(\sum_{m \in M} \max_{s \in S}(P_{jms})\right) & Rd/Dd = 1
\end{cases}
    \label{eq:maxTardiness}
\end{equation}

Maximum Tardiness (Equation \ref{eq:maxTardiness}) represents the maximum possible delay in a solution. If there are no release or due date constraints ($Rd/Dd = 0$), it is set to -1. Otherwise, it sums the maximum processing times, indicating the worst-case delay scenario.

\begin{equation}
\begin{cases}
-1 & Rd/Dd = 0 \\ \\
\frac{\displaystyle\sum_{j \in J} \frac{Dd_j - Rd_j}{\sum_{m \in M} P_{jm}}}{|J|} & Rd/Dd = 1 \\ \\
\frac{\displaystyle\sum_{j \in J} \sum_{m \in M} \frac{Dd_{jm} - Rd_{jm}}{P_{jm}}}{|J| \cdot |M|} & Rd/Dd = 2
\end{cases}
\label{eq:timeWindow}
\end{equation}

Time-Window (Equation \ref{eq:timeWindow}) represents the number of times a job or operation can be performed within its time window. This metric varies based on the type of release and due date constraints: no constraints, job-level constraints, or operation-level constraints, indicating flexibility in scheduling.

\begin{figure*}[t]
\begin{equation}
    \begin{cases}
        -1 & Rd/Dd = 0 \\ \\
        \frac{\displaystyle\sum_{\substack{j_1, j_2 \in J \\ j_1 \neq j_2}} \frac{\max(0, \min(Dd_{j_1}, Dd_{j_2}) - \max(Rd_{j_1}, Rd_{j_2}))}{Dd_{j_1} - Rd_{j_1}}}{|J| \cdot (|J| - 1)} & Rd/Dd = 1 \\ \\
        \frac{\displaystyle\sum_{\substack{j_1, j_2 \in J \\ j_1 \neq j_2}} \sum_{m \in M} \frac{\max(0, \min(Dd_{j_{1}m}, Dd_{j_{2}m}) - \max(Rd_{j_{1}m}, Rd_{j_{2}m}))}{Dd_{j_{1}m} - Rd_{j_{1}m}}}{|J| \cdot (|J| - 1) \cdot |M|} & Rd/Dd = 2
    \end{cases}
    \label{eq:overlap}
\end{equation}
\end{figure*}


Overlap (Equation \ref{eq:overlap}) represents the degree of overlap between the time windows of the jobs or operations. This metric assesses how much the scheduling windows for different jobs or operations coincide, which impacts the complexity and difficulty of scheduling.

\subsection{Machine Learning Models}

Once the instances have been vectorized and solved, a tabular data set is constructed with the characteristics of each instance and the solver that has found the best solution, that is, the one that has obtained the lowest value of the objective function.

This dataset has been separated into two subsets, a training subset with a size of 80\% and a test subset with the remaining 20\%. In addition, it has been ensured that the same number in proportion of instances exists in the two subsets. 

The training dataset has been used to validate different models using five-fold cross-validation. The validated models are the following:

\begin{itemize}
    \item Logistic Regression: This is a statistical method for analyzing a dataset in which one or more independent variables determine an outcome. The outcome is measured with a dichotomous variable (i.e., two possible outcomes). Logistic regression is particularly useful for binary classification problems and provides insights into the relationships between the variables and the probability of the outcomes.
    
    \item Decision Tree: This is a decision support tool that uses a tree-like model of decisions and their possible consequences, including chance event outcomes, resource costs, and utility. A decision tree is built by splitting the dataset into subsets based on the value of input features, with the goal of making the most informative splits. This method is easy to interpret and visualize, making it useful for understanding the structure of the data.

    \item Gaussian Naive Bayes: This is a probabilistic classifier based on Bayes' theorem, with the assumption that the features are independent given the class label and that they follow a Gaussian distribution. Despite its simplicity, Gaussian Naive Bayes can perform well in various situations, especially when the assumption of independence roughly holds true.

    \item K-Nearest Neighbors (KNN): This is a non-parametric method used for classification and regression. For classification, the input consists of the $k$ closest training examples in the feature space, and the output is a class membership. The object is assigned to the class most common among its $k$ nearest neighbors.

    \item Random Forest: This is an ensemble learning method for classification and regression that constructs multiple decision trees during training and outputs the mode of the classes (classification) or mean prediction (regression) of the individual trees. Random forests improve the predictive accuracy and control over-fitting by averaging multiple trees, reducing the model's variance.

    \item XGBoost \cite{Chen_2016}: This is an optimized distributed gradient boosting library designed to be highly efficient, flexible, and portable. It implements machine learning algorithms under the gradient boosting framework, which builds models in a stage-wise fashion and generalizes them by optimizing for a differentiable loss function. XGBoost is known for its speed and performance, making it a popular choice for structured/tabular data.

    \item Multi-Layer Perceptron (MLP): This is a class of feedforward artificial neural networks that consist of at least three layers of nodes: an input layer, a hidden layer, and an output layer. Except for the input nodes, each node (or neuron) uses a nonlinear activation function. MLPs are capable of learning complex mappings from inputs to outputs and are trained using backpropagation.
\end{itemize}

\section{Evaluation}
All experiments were conducted on a system equipped with an Intel 3.60 GHz 12th generation Core i7 CPU and 64 GB of RAM. The implementation was developed in Python 3.11. Well-known solvers such as GUROBI \cite{GUROBI}, CPLEX \cite{CPLEX}, and GECODE \cite{GECODE}, which are implemented on Minizinc, were utilized.

To evaluate the quality of the solutions obtained, the mono-objective function shown in Equation \ref{eq:obmulti} is used to compare the best solutions from the solvers. Other important results, such as the average objective function, solving time, optimum, satisfaction rate, and the number of unsolved solutions, are presented above.

\subsection{Instances}
Instance creation is one of the most important aspects of evaluation as it allows a specific number of instances to be configured to ensure the most comprehensive evaluation possible, taking into account all possible combinations the problem may encounter in real-life scenarios.

The JSP Benchmark used for testing is composed of the number of jobs ($J$) and machines ($M$) to determine each job's tasks. These variables can take any natural number. In this test set, the set $\{5,10,20,25,50,100\}$ is considered for $J$. The release and due date can take values $\{0, 1, 2\}$, speed scaling can take values $\{1, 3, 5\}$, and statistical distributions considered are $\{\textit{uniform},\textit{normal},\textit{exponential}\}$. For each configuration, 10 instances are generated with different seeds to ensure substantial variation between them. Therefore, a total of $6 (J) \times 6 (M) \times 3 (rddd) \times 3 (ss) \times 3 (dist) \times 10 (Q) = 9720$ instances are obtained.
\begin{figure}[hbtp]
    \centering
     \includegraphics[width=0.75\linewidth]{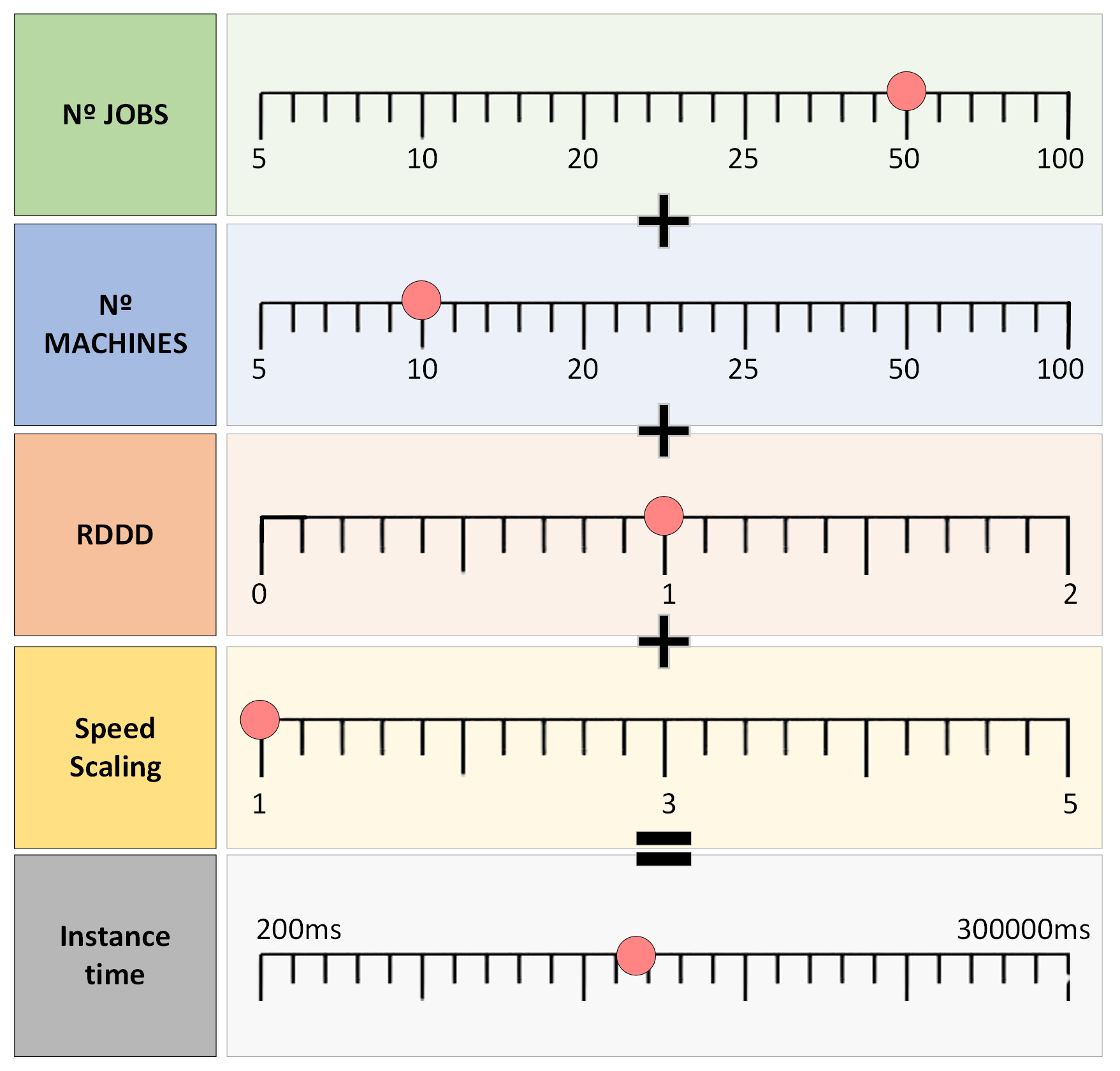}
    \caption{Distribution of timeout and relationship between time and energy.}
    \label{fig:timeinstance}
\end{figure}

The time allocated for resolving each instance depends on the specific characteristics of the problem. An example of this allocation is illustrated in Figure \ref{fig:timeinstance}. In this example, an instance with 50 jobs, 10 machines, no Release Date or Due Date, and a single speed per machine is allocated 149 seconds. The principle is that if the maximum allocation time for an instance is 300,000 milliseconds, each characteristic's impact on the allocation should be equivalent. Therefore, each characteristic contributes at most $300,000 / 4 = 75,000$ milliseconds. In this manner, for the given example, if the Release Date and Due Date are assigned at the operation level (RDDD = 2), they contribute 75,000 milliseconds. If they are absent (RDDD = 0), they contribute 50 milliseconds. When assigned at the job level, the contribution is determined by exponential interpolation between these two cases.

\subsection{Results}
Upon defining the problem instances and setting appropriate search time limits for each solver, the focus shifted toward analyzing and interpreting the outcomes. This involved evaluating the efficacy of the solvers employed, assessing solution quality, and considering the broader implications within the problem domain.

\begin{figure}[hbtp]
    \centering
     \includegraphics[width=\linewidth]{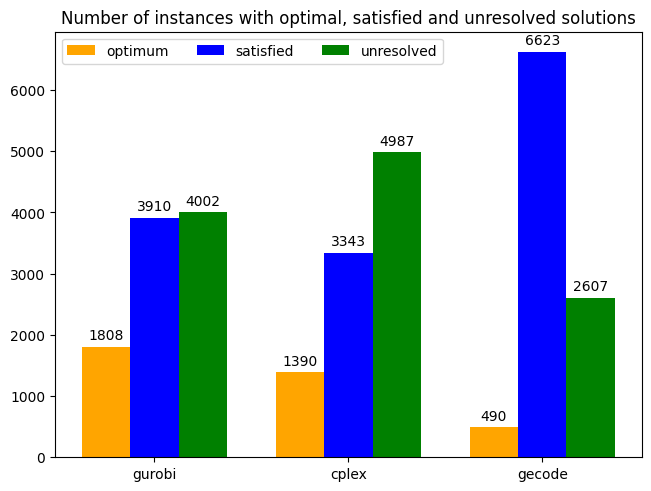}
    \caption{Quantity of optimum, satisfied and unresolved instances by each solver. }
    \label{fig:optsatun}
\end{figure}

Figure \ref{fig:optsatun} illustrates the distribution of solved instances among GUROBI, CPLEX, and GECODE across three categories: optimal (best solution), satisfied (feasible but not optimal), and unresolved (not solved).

GUROBI emerged as the top performer overall, solving the highest number of optimal solutions and consistently demonstrating its ability to find acceptable solutions even when optimal ones were unfeasible. This underscores GUROBI's robust capability in efficiently managing a diverse array of problem types. Conversely, GECODE excelled in finding feasible solutions, significantly outperforming other solvers in achieving satisfactory solutions. Moreover, GECODE showed the fewest instances left unresolved, highlighting its reliability in tackling complex problems without abandoning them.

In contrast, CPLEX, while proficient, faced challenges with more complex problem instances, leading to a higher incidence of unresolved cases. Although it achieved reasonable numbers of optimal and satisfactory solutions, its performance consistency was observed to be less reliable compared to GUROBI and GECODE.

\begin{table*}[hbtp]
\resizebox{0.75\linewidth}{!}{%
\begin{tabular}{|c|c|ccc|ccc|}
\hline
\multirow{2}{*}{\textbf{Jobs}} &
  \multirow{2}{*}{\textbf{Machines}} &
  \multicolumn{3}{c|}{\textbf{Solve Time}} &
  \multicolumn{3}{c|}{\textbf{Objective}} \\ \cline{3-8} 
 &
   &
  \multicolumn{1}{c|}{\textbf{GUROBI}} &
  \multicolumn{1}{c|}{\textbf{CPLEX}} &
  \textbf{Gecode} &
  \multicolumn{1}{c|}{\textbf{GUROBI}} &
  \multicolumn{1}{c|}{\textbf{CPLEX}} &
  \textbf{Gecode} \\ \hline
\multirow{6}{*}{5}   & 5   & \textbf{6499.47}   & 9393.54            & 76754.74           & 0.58053          & \textbf{0.57924} & 0.7748           \\
                     & 10  & \textbf{22177.73}  & 35193.18           & 77122.47           & 0.46884          & \textbf{0.46842} & 0.64781          \\
                     & 20  & \textbf{50222.29}  & 57898.63           & 81637.16           & \textbf{0.42377} & 0.42422          & 0.58867          \\
                     & 25  & \textbf{55920.19}  & 59732.08           & 79771.76           & \textbf{0.41208} & 0.41279          & 0.5677           \\
                     & 50  & \textbf{69226.88}  & 71453.81           & 95645.56           & \textbf{0.39632} & 0.40691          & 0.53307          \\
                     & 100 & \textbf{110177.22} & 123730.15          & 154180.72          & \textbf{0.38944} & 0.42882          & 0.51176          \\ \hline
\multirow{6}{*}{10}  & 5   & \textbf{72263.13}  & 76261.31           & 85388.49           & \textbf{0.56284} & 0.56433          & 0.77965          \\
                     & 10  & \textbf{69831.23}  & 71671.93           & 86539.69           & \textbf{0.44524} & 0.4482           & 0.64365          \\
                     & 20  & \textbf{71619.72}  & 74610.4            & 89860.43           & \textbf{0.38908} & 0.39532          & 0.5713           \\
                     & 25  & \textbf{73930.71}  & 78202.9            & 91969.19           & \textbf{0.3798}  & 0.39247          & 0.54207          \\
                     & 50  & \textbf{89935.3}   & 95355.06           & 109638.51          & \textbf{0.36431} & 0.38019          & 0.49456          \\
                     & 100 & \textbf{146282.37} & 152079.82          & 168202.69          & 0.35029          & \textbf{0.30281} & 0.35159          \\ \hline
\multirow{6}{*}{20}  & 5   & \textbf{80190.01}  & 88650.74           & 89079.98           & \textbf{0.65727} & 0.67151          & 0.83772          \\
                     & 10  & \textbf{80369.98}  & 89368.09           & 89895.89           & \textbf{0.48249} & 0.48393          & 0.71326          \\
                     & 20  & 91070.63           & \textbf{83648.86}  & 92953.44           & 0.40615          & \textbf{0.36374} & 0.60325          \\
                     & 25  & 92555.52           & \textbf{86596.51}  & 96516.25           & 0.38184          & \textbf{0.3135}  & 0.58104          \\
                     & 50  & 109441.69          & 97752.14           & \textbf{78306.1}   & 0.35987          & 0.24087          & \textbf{0.20729} \\
                     & 100 & 164169.07          & 161843.67          & \textbf{112923.84} & 0.29168          & 0.06245          & \textbf{0.00733} \\ \hline
\multirow{6}{*}{25}  & 5   & \textbf{73156.78}  & 86322.15           & 91381.59           & 0.64018          & \textbf{0.63401} & 0.84305          \\
                     & 10  & \textbf{78504.99}  & 83658.47           & 92142.26           & 0.57263          & \textbf{0.46831} & 0.73123          \\
                     & 20  & 99400.75           & \textbf{83989.35}  & 95774.25           & 0.46229          & \textbf{0.36985} & 0.62663          \\
                     & 25  & 104469.18          & \textbf{76519.47}  & 102202.35          & 0.42563          & \textbf{0.30161} & 0.61462          \\
                     & 50  & 118524.24          & 127129.68          & \textbf{59800.26}  & 0.31818          & 0.21552          & \textbf{0.13974} \\
                     & 100 & 168354.77          & \textit{Timeout}   & \textbf{113885.23} & 0.28319          & \textit{Timeout} & \textbf{0.01021} \\ \hline
\multirow{6}{*}{50}  & 5   & 101811.91          & \textbf{86469.46}  & 110038.97          & 0.70258          & \textbf{0.49251} & 0.86043          \\
                     & 10  & 115365.19          & \textbf{71684.29}  & 110542.27          & 0.56627          & \textbf{0.21168} & 0.75717          \\
                     & 20  & 108631.5           & 117752.67          & \textbf{78930.63}  & 0.37003          & \textbf{0.12673} & 0.66168          \\
                     & 25  & 105251.06          & 120220             & \textbf{72639.27}  & 0.43698          & \textbf{0.16123} & 0.57382          \\
                     & 50  & 136252.05          & \textit{Timeout}   & \textbf{59343.32}  & 0.47363          & \textit{Timeout} & \textbf{0.14455} \\
                     & 100 & 153826.43          & \textit{Timeout}   & \textbf{122941.2}  & 0.91743          & \textit{Timeout} & \textbf{0.04168} \\ \hline
\multirow{6}{*}{100} & 5   & \textit{Timeout}   & \textbf{114685.56} & 183151.94          & \textit{Timeout} & \textbf{0.32913} & 0.86754          \\
                     & 10  & \textit{Timeout}   & \textit{Timeout}   & \textbf{180057.66} & \textit{Timeout} & \textit{Timeout} & \textbf{0.77187} \\
                     & 20  & \textit{Timeout}   & \textit{Timeout}   & \textbf{139246.52} & \textit{Timeout} & \textit{Timeout} & \textbf{0.68049} \\
                     & 25  & \textit{Timeout}   & \textit{Timeout}   & \textbf{122740.76} & \textit{Timeout} & \textit{Timeout} & \textbf{0.59173} \\
                     & 50  & \textit{Timeout}   & \textit{Timeout}   & \textbf{116506.13} & \textit{Timeout} & \textit{Timeout} & \textbf{0.4602}  \\
                     & 100 & \textit{Timeout}   & \textit{Timeout}   & \textbf{157021.85} & \textit{Timeout} & \textit{Timeout} & \textbf{0.13594} \\ \hline
\end{tabular}%
}
\caption{Comparison of mean resolution time and mean objective function obtained with different solvers.}
\label{tab:resume}
\end{table*}

Table \ref{tab:resume} compares solution times and objective function values from GUROBI, CPLEX, and GECODE across different job and machine configurations. This analysis reveals insights into each solver's performance characteristics, highlighting strengths and limitations in solving optimization problems.

For 5 to 20 jobs, GUROBI consistently shows shorter solution times and competitive objective values compared to CPLEX and GECODE. Its efficiency and precision make it highly effective in simpler problem instances.

In medium-sized scenarios (20 to 50 jobs), GUROBI maintains an edge, particularly with fewer machines, though CPLEX occasionally performs better in specific configurations. GUROBI generally achieves superior objective function values in varied problem setups.

In complex cases (50 to 100 jobs), GECODE demonstrates exceptional scalability despite encountering timeouts in some instances. GUROBI and CPLEX struggle more often with timeouts as problem size and complexity increase, yet GUROBI often maintains competitive objective values.

These insights underscore the importance of selecting solvers based on problem specifics. GUROBI excels in smaller to medium-sized instances, balancing efficiency and high-quality solutions. CPLEX performs well in certain medium-sized setups but faces scalability challenges. GECODE shines in complex problems, offering robust scalability and reliability despite occasional computational hurdles. These findings aid practitioners in optimizing solver choices and considering trade-offs between solution quality, efficiency, and problem complexity.
\subsection{Complexity analysis}

Observing the results obtained by the methods used, a relationship is observed between the parameters employed and the complexity of the instances. This part of the study focuses on the in-depth analysis of each parameter to observe its contribution to the overall complexity of the instances.

\begin{figure}[hbtp]
     \centering
     \begin{subfigure}{0.49\textwidth}
         \centering
          \includegraphics[width=\textwidth]{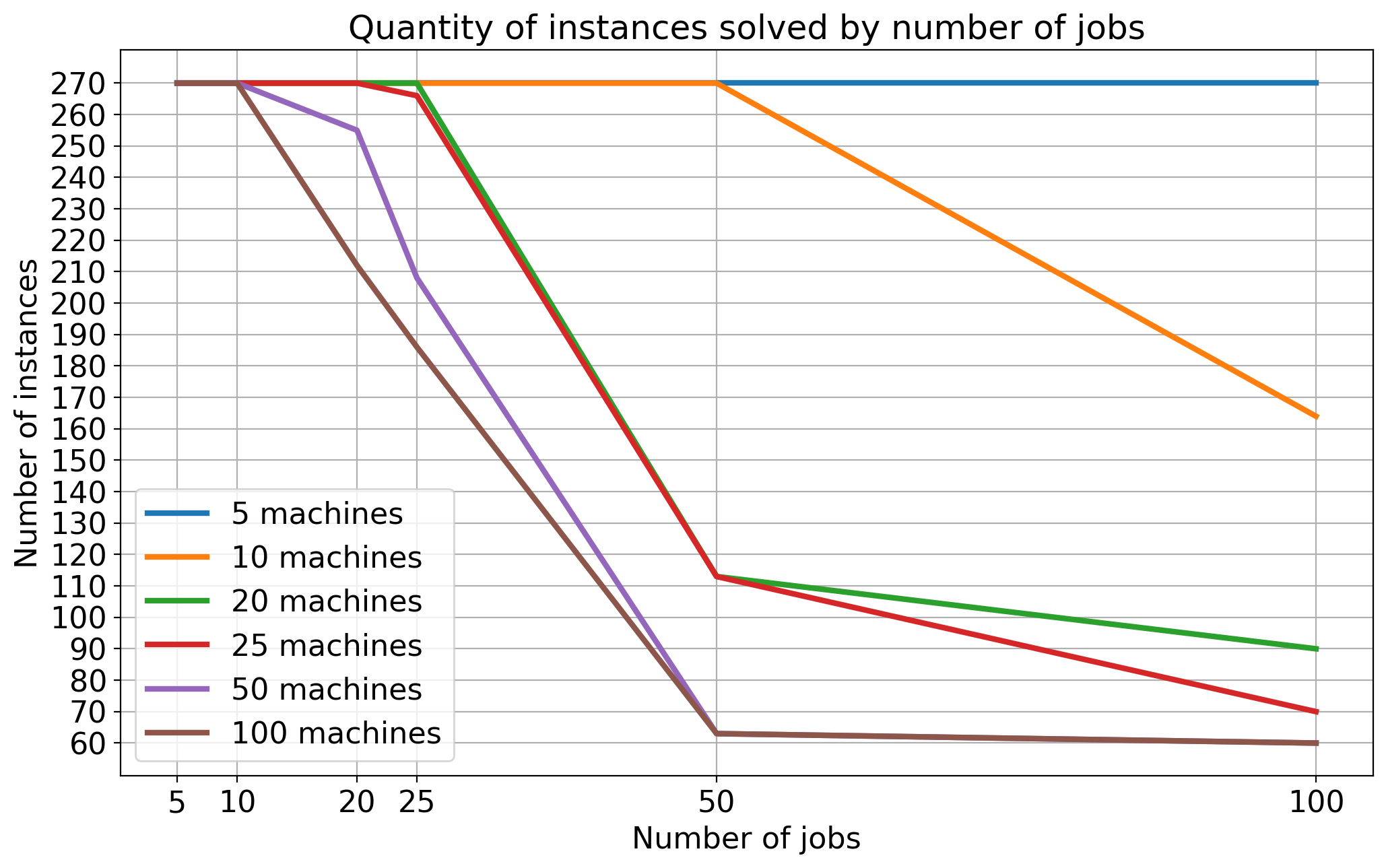}
         \caption{by number of machines}
         \label{fig:jobs-machines}
     \end{subfigure}
     \begin{subfigure}{0.49\textwidth}
         \centering
          \includegraphics[width=\textwidth]{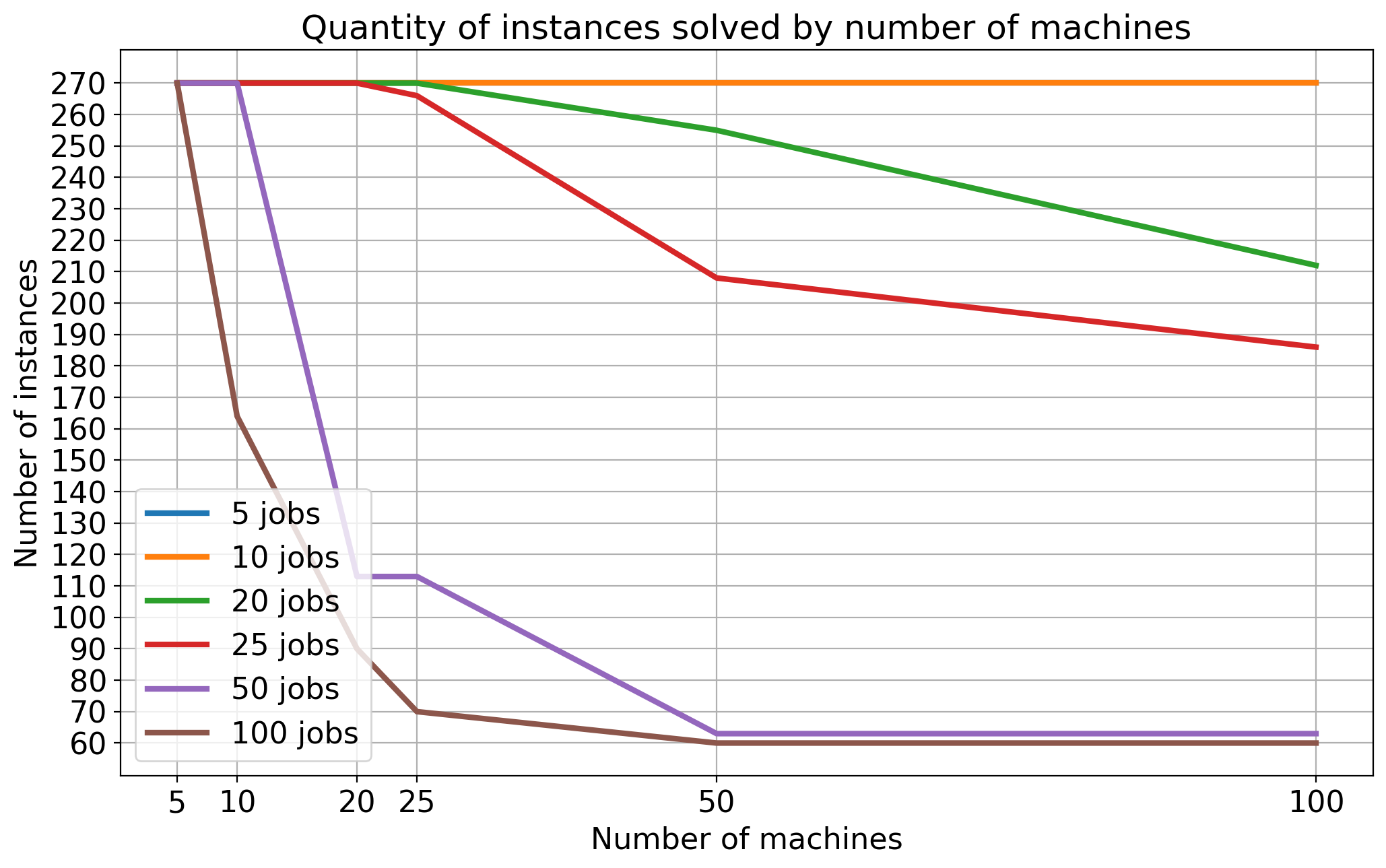}
         \caption{by number of jobs}
         \label{fig:machines-jobs}
     \end{subfigure}
    \caption{Number of instances solved}
    \label{fig:instancesolved}
    
\end{figure}

To delve deeper into the data presented in Table \ref{tab:resume}, Figure \ref{fig:instancesolved} provides a general overview of the number of solved instances. Organizing the data by the number of machines, as shown in subfigure \ref{fig:jobs-machines}, it is evident that as the number of machines increases, the number of solved instances progressively decreases except for the case of 5 machines, where all instances are solved for all possible job configurations. Looking at subfigure \ref{fig:machines-jobs}, which is organized by the number of jobs for all possible machine configurations, it can be seen that all instances are solved for 5 and 10 jobs, but there is a notable decrease for the rest. Focusing on the set of 20 and 25 jobs, it can be observed that there is a slight decrease from 25 machines onwards; later, the reasons for this are analyzed. For instances with 50 and 100 jobs, the decrease in the number of solved instances is exponential. Only all instances are solved for a configuration of 5 machines for 100 jobs and 5 and 10 machines for 50 jobs. This figure illustrates how the number of jobs and machines affects the possibility of obtaining a solution to the problem at hand.

\subsection{Algorithm selector results}

\begin{table}[hbtp]
    \resizebox{\linewidth}{!}{
        \begin{tabular}{|l|c|}
            \hline
            \textbf{Model}           & \textbf{Accuracy (\%)} \\ \hline
            Logistic Regression      & 76.08 \\ \hline
            Gaussian Naive Bayes     & 48.93 \\ \hline
            Decision Tree            & 79.48 \\ \hline
            $K$-Nearest Neighbors    & 78.34 \\ \hline
            Random Forest            & 82.87 \\ \hline
            XGBoost                  & \textbf{83.26} \\ \hline
            MLP                      & 82.81 \\ \hline
        \end{tabular}%
    }
\caption{Table showing the training results of the different models}
\label{tab:valrecommender}
\end{table}

Table \ref{tab:valrecommender} shows the validation results of the tested models. As can be seen, XGBoost is the model with the best validation accuracy. Training this model with the total training data set and testing it with the test set finally yields an accuracy of 84.51\%. This indicates that XGBoost performs well during the validation phase and generalizes effectively to unseen data. The high accuracy suggests that XGBoost's ensemble learning approach, which combines multiple decision trees to improve performance, is particularly well-suited to this dataset. Moreover, the performance difference between XGBoost and other models like Random Forest and MLP, which also showed strong results, 

\begin{figure}[hbtp]
    \centering
     \includegraphics[width=0.75\linewidth]{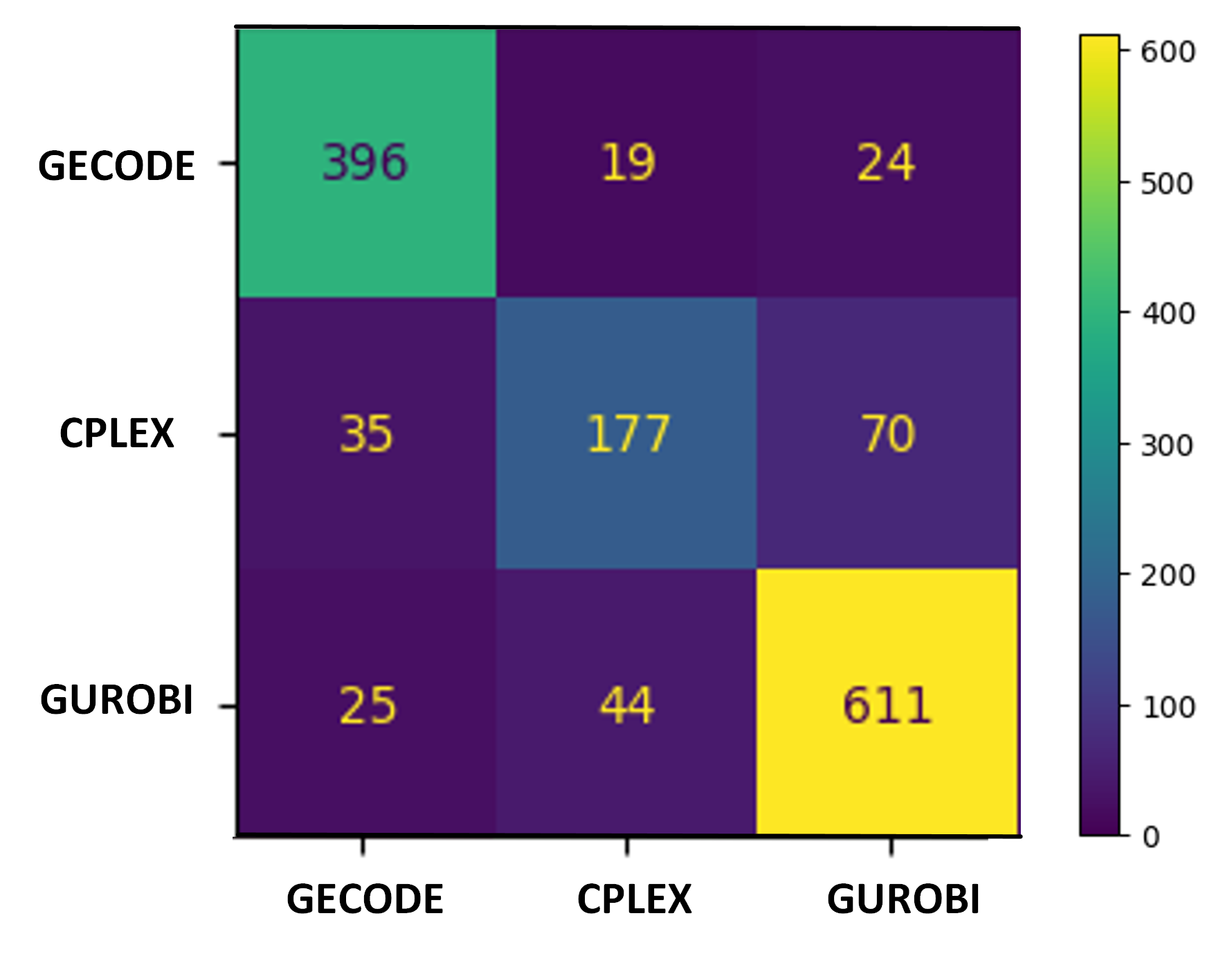}
    \caption{Confusion matrix for the algorithm selector predictions (true algorithms on the y-axis, predicted algorithms on the x-axis).}
    \label{fig:confusion}
\end{figure}

The confusion matrix in Figure \ref{fig:confusion} presents the algorithm selector's classification results. Each cell value represents the number of instances where the algorithm selector predicted the algorithm in the corresponding column for a problem best solved by the algorithm in the corresponding row. For instance, the selector correctly identified GUROBI for 611 out of the total instances where GUROBI was the best choice. Similarly, GECODE was correctly identified 396 times. However, there are misclassifications, such as predicting GUROBI when CPLEX was optimal, which occurred 70 times.

An in-depth analysis of the precision and recall metrics provides further insights into the performance of the algorithm selector. GECODE achieved a precision of 90.20\%, indicating that 90.20\% of the instances predicted as GECODE were correctly identified. Its recall was 86.84\%, signifying that 86.84\% of the actual GECODE instances were correctly detected. This high precision and recall demonstrate the algorithm selector's robustness in identifying GECODE instances accurately.

On the other hand, CPLEX showed a precision of 62.76\%, meaning that only 62.76\% of the predictions for CPLEX were accurate, and a recall of 73.75\%, which indicates that 73.75\% of the actual CPLEX instances were correctly classified. The lower precision for CPLEX suggests a higher rate of false positives, which could imply that the algorithm selector often misclassifies other algorithms as CPLEX.

For GUROBI, the precision was 89.85\%, reflecting that 89.85\% of the GUROBI predictions were correct, and the recall was 86.66\%, meaning that 86.66\% of the actual GUROBI instances were identified correctly. These values indicate a strong performance, similar to GECODE, highlighting the selector's efficiency in recognizing GUROBI accurately.

These metrics, precision, and recall, are crucial for evaluating the algorithm selector's effectiveness, as they provide a more comprehensive understanding of its performance beyond simple accuracy. They highlight the selector's strengths in accurately identifying certain algorithms while also pointing out areas where misclassification occurs, thus providing a clear direction for further improvements.

\section{Conclusions}

This study explores the complexities of JSP, emphasizing its NP-completeness and diverse optimization goals such as makespan, energy consumption, and tardiness. The problem presents significant computational challenges due to its combinatorial nature, making timely optimal solutions crucial in operations research and manufacturing.

An innovative aspect of this research is integrating machine learning techniques to enhance algorithm selection for JSP instances. By extracting comprehensive features like job and machine characteristics, release dates, and energy requirements, models such as XGBoost and Random Forest were effectively trained. These models accurately recommend suitable solvers like GUROBI, CPLEX, and GECODE, streamlining decision-making for solving diverse and complex scheduling problems.

GUROBI proved particularly efficient for smaller to medium-sized instances, consistently delivering optimal and satisfactory solutions across different configurations. Meanwhile, GECODE demonstrated robust scalability, excelling in complex scenarios despite occasional computational challenges. This analysis underscores the importance of selecting solvers based on specific problem parameters to optimize solution quality and computational efficiency.

Looking ahead, the study suggests refining feature extraction methodologies to enhance the algorithm selector's accuracy across a broader range of JSP scenarios. Advancements in solver performance under varying constraints promise to expand the practical utility of scheduling optimization tools in real-world manufacturing settings, emphasizing efficiency and sustainability.

Although the results obtained are not as high as those reported in the literature, it should be noted that the energy-aware JSP constitutes a more complex problem compared to the standard JSP and flexible JSP found in the literature. Furthermore, this study uses a smaller set of features than those used in other studies, yet the accuracy achieved is not significantly lower.

In conclusion, this work advances both academic understanding and practical applications by integrating traditional optimization techniques with modern machine-learning approaches. It offers tools that can significantly benefit research and industrial practices, addressing contemporary challenges in operations management and manufacturing logistics.

\section{Acknowledgments}
The authors gratefully acknowledge the financial support of the European Social Fund (Investing In Your Future), the Spanish Ministry of Science (project  PID2021-125919NB-I00).

\bibliography{bib}

\end{document}